%% file: bmvc_final.tex
\documentclass{bmvc2k}
\usepackage{amsfonts}
\usepackage{amssymb}
\usepackage{graphicx}
\usepackage{booktabs}
\usepackage{array}
\usepackage[table]{xcolor}
\usepackage{colortbl}
\newcommand{\best}{\cellcolor{red!50}}
\newcommand{\sbest}{\cellcolor{orange!50}}
\newcommand{\tbest}{\cellcolor{yellow!50}}
\title{MonoGSDF: Exploring Monocular Geometric Cues for Gaussian Splatting-Guided Implicit Surface Reconstruction}

\addauthor{Kunyi Li}{kunyi.li@tum.de}{1, 4}
\addauthor{Michael Niemeyer}{mniemeyer@google.com}{2}
\addauthor{Zeyu Chen}{chen-zy22@mails.tsinghua.edu.cn}{3}
\addauthor{Nassir Navab}{nassir.navab@tum.de}{1, 4}
\addauthor{Federico Tombari}{tombari@in.tum.de}{1, 2}

\addinstitution{
 Technical University of Munich \\
 Munich, Germany
}
\addinstitution{
 Google\\
 Zurich, Switzerland
}
\addinstitution{
 Tsinghua University\\
 Beijing, China
}
\addinstitution{
 Munich Center for Machine Learning\\
 Munich, Germany
}

\runninghead{K.Li, M.Niemeyer, Z.Chen, N.Navab, F.Tombari}{MonoGSDF}

\begin{document}

\maketitle

\input{sec/0_abstract}
\input{sec/1_intro}
\input{sec/2_related}
\input{sec/3_method}
\input{sec/4_experiments}
\input{sec/5_conclusion}
{
    % \small
    % \bibliographystyle{ieeenat_fullname}
    \bibliography{main}
}
\newpage
\ 
\newpage
\input{sec/X_suppl}

\end{document}

%% file: sec/0_abstract.tex
\begin{abstract}
Accurate meshing from images remains a key challenge in 3D vision. While state-of-the-art 3D Gaussian Splatting (3DGS) methods excel at synthesizing photorealistic novel views through rasterization-based rendering, their reliance on sparse, explicit primitives severely limits their ability to recover watertight and topologically consistent 3D surfaces.
We introduce MonoGSDF, a novel method that couples Gaussian-based primitives with a neural Signed Distance Field (SDF) for high-quality reconstruction with monocular RGB images as input. During training, the SDF guides Gaussians' spatial distribution, while at inference, Gaussians serve as priors to reconstruct surfaces, eliminating the need for memory-intensive Marching Cubes.
To handle arbitrary-scale scenes, we propose a scaling strategy for robust generalization. A multi-resolution training scheme further refines details and monocular geometric cues from off-the-shelf estimators enhance reconstruction quality. Experiments on real-world datasets show MonoGSDF outperforms prior methods while maintaining efficiency.
\end{abstract}

%% file: sec/1_intro.tex
\section{Introduction}
\label{sec:intro}
\begin{figure}[t]
\includegraphics[width=12.5cm]{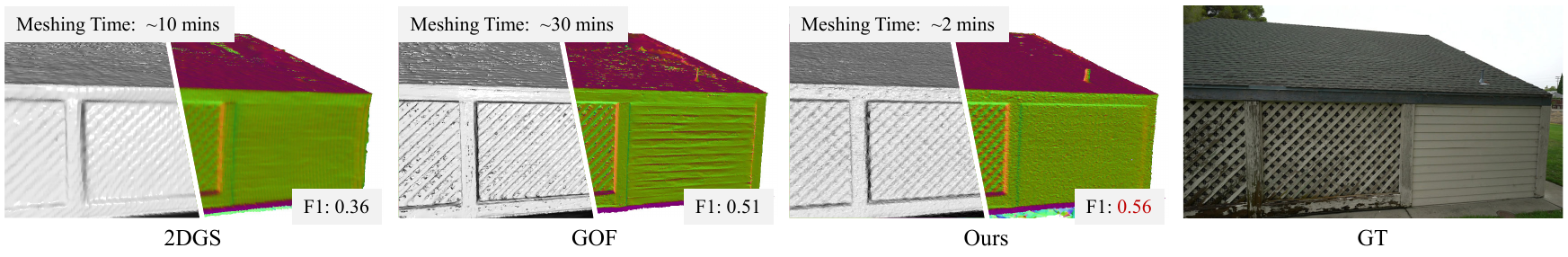}
\centering
\vspace{-0.2cm}
\caption{\textbf{MonoGSDF.} We show the reconstructed mesh. Compared to 2DGS \cite{huang20242d} and GOF \cite{yu2024gaussian}, ours achieves higher F1 scores ($\uparrow$) and reconstructs smooth surfaces with fine details.} 
\label{fig:teaser}
\vspace{-0.5cm}
\end{figure}

3D Gaussian Splatting (3DGS) \cite{kerbl20233d} has emerged as a state-of-the-art method for high-quality novel view synthesis by leveraging the rasterization pipeline and representing 3D scenes using points characterized by Gaussian functions. However, 3DGS remains an explicit and discrete representation, posing challenges for accurate 3D surface reconstruction and meshing. This limitation is particularly important for applications such as geometry editing \cite{yang2024deformable, chen2024gaussianeditor, wang2024gaussianeditor}, 3D animation \cite{tang2023dreamgaussian, yuan2024gavatar, qian20243dgs}, and robotics \cite{zhou2024drivinggaussian, keetha2024splatam, yugay2023gaussian}. The main challenge is that 3DGS models geometry as a discrete, unstructured point set. Because 3D Gaussians are optimized for rendering quality instead of geometric consistency, post-processing methods such as Poisson surface reconstruction \cite{kazhdan2006poisson} struggle to recover accurate 3D surfaces.

These challenges have motivated recent investigations exploring how 3DGS can be used for high-quality surface reconstruction while maintaining its rendering speed and efficiency. Methods like 2DGS \cite{huang20242d} address this by aligning 2D Gaussians to surfaces, although their depth map-based mesh extraction \cite{curless1996volumetric} struggles with thin structures and unbounded scenes due to resolution and resource limitations. Methods such as SuGaR \cite{guedon2024sugar} attempt to address these limitations by aligning 3D Gaussians with mesh faces. Nevertheless their reliance on Poisson reconstruction \cite{kazhdan2006poisson} still fundamentally limits reconstruction and meshing quality due to the inherent geometric inaccuracies of Gaussian representations. Although recent methods \cite{yu2024gsdf, lyu20243dgsr, zhang2025neural} have integrated SDF fields into Gaussian training for surface reconstruction, their incomplete exploration of the Gaussian-SDF relationship results in over-smoothed surfaces and limits their application within object-level reconstruction.

To bridge the gap between explicit Gaussian representations and implicit surface reconstruction while fully leveraging their complementary advantages, we present MonoGSDF, a unified framework that establishes a robust connection between 3D Gaussians and SDF-based surface modeling for high-fidelity reconstruction with only monocular RGB images as input. Our \textbf{contributions} are:
1. a Gaussian-based implicit surface reconstruction pipeline for monocular images, where the SDF guides Gaussians to better align near the surface during training and Gaussians act as priors for fast mesh extraction;
2. a simple yet efficient SDF-to-Opacity function that bridges implicit SDF with explicit Gaussians, coupled with a normalization strategy that maps unbounded 3D coordinates to a (0,1) range with minimal distortion;
3. a multi-resolution strategy based on wavelet transforms progressively refines details, while off-the-shelf geometric cues further enhance reconstruction quality.
Experiments on real-world datasets demonstrate that our method delivers high-quality reconstructions and outperforms baselines by 13\% in terms of Chamfer distance 
%on DTU \cite{jensen2014large} 
while preserving the view synthesis quality. 
%Note that our method is also easily adaptable to any existing Gaussian rasterizer.

%% file: sec/2_related.tex
\section{Related Works}
\label{sec:related}

\subsection{View Synthesis and Gaussian Splatting}
Neural Radiance Field (NeRF) \cite{mildenhall2021nerf} utilize multi-layer perceptions (MLP) as scene representation to predict geometry and view-dependent appearance. The MLP is optimized via a photometric loss through volume rendering. Subsequent methods have focused on optimizing NeRF’s training and expressiveness using grid representations \cite{muller2022instant}, improving rendering speed \cite{sun2022direct, zhang2020nerf} and scaling to unbounded scenes \cite{barron2022mip}. However, volume rendering typically requires substantial computational resources and extensive training durations.
3D Gaussian Splatting (3DGS) \cite{kerbl20233d} has emerged as an efficient approach for real-time view synthesis through differentiable Gaussian functions. Subsequent works have enhanced rendering quality via anti-aliasing techniques \cite{yu2024mip, yan2024multi, song2024sa} and improved rendering speed through Gaussian density control \cite{yang2024spectrally} and radiance field priors \cite{liu2025mvsgaussian, niemeyer2024radsplat}. Geometry-focused approaches like DNGaussian \cite{li2024dngaussian} address sparse view degradation, while GeoGaussian \cite{li2025geogaussian} preserves non-textured regions. Instantsplat \cite{fan2024instantsplat} accelerates sparse view training using Dust3r \cite{wang2024dust3r} initialization, and Scaffold-GS \cite{lu2024scaffold} combines implicit-explicit representations. However, these methods primarily focus on appearance quality rather than underlying geometry, limiting their application to view synthesis.

\subsection{Surface Reconstruction from Gaussians}
Due to the discrete and unstructured nature of 3DGS, along with supervision being limited to RGB images during training, existing methods often struggle to accurately capture scene geometry, making surface reconstruction particularly challenging. SuGaR \cite{guedon2024sugar} addresses this by constructing a density field from Gaussians and extracting meshes via level-set searching. However, it is computationally expensive and struggles to reconstruct large, smooth surfaces like floors and walls.  
Other approaches \cite{chen2024pgsr, chen2024vcr, zhang2024rade, turkulainen2024dn, wolf2024surface} incorporate depth or normal estimators as priors to supervise Gaussians. 2DGS \cite{huang20242d} and GSurfels \cite{dai2024high} improves geometric alignment by flattening 3D Gaussians into 2D disks or surfels, enabling better surface representation. For surface reconstruction, they rely on either multi-view depth maps for TSDF fusion or Poisson reconstruction. However, TSDF fusion is constrained by fixed resolution and high memory usage for large scenes. To handle unbounded scenes, it requires contracting coordinates, introducing unnecessary distortions. Meanwhile, Poisson reconstruction often produces noisy surfaces due to inaccuracies in Gaussian geometry distribution.  
Instead of refining depth and normal maps, some methods \cite{yu2024gsdf, zhang2025neural, lyu20243dgsr, baixin2024gsurf} integrate Signed Distance Field (SDF). 3DGSR \cite{lyu20243dgsr} and GSDF \cite{yu2024gsdf} associate Gaussians with SDF via an additional branch that volume-renders depth and normal supervised by rendered Gaussian depth and normal. However, this approach is inefficient and fails to fully exploit the relationship between Gaussians and SDF. GS-Pull \cite{zhang2025neural} and GSurf \cite{baixin2024gsurf} improve upon this by leveraging SDF gradients to better align Gaussians with surfaces, but they are either restricted to object-level reconstruction or produce overly smooth results.
Gaussian Opacity Fields (GOF) \cite{yu2024gaussian} approximates minimum accumulated alpha values from all views via ray tracing to construct an opacity field, followed by surface extraction using Marching Tetrahedra \cite{kulhanek2023tetra}. However, Gaussian distributions remain inconsistent and unstructured relative to scene surfaces.  
Our method leverages SDF to guide Gaussians to align closer to the surface during training. At inference, Gaussians serve as primitives to efficiently constrain zero-level set extraction, avoiding redundant free-space searches while achieving accurate reconstruction.

%% file: sec/3_method.tex
\section{Method}
\label{sec:method}

\subsection{Preliminaries}
\label{subsec:pre}
\paragraph{3D Gaussian Splatting.}
3D Gaussian Splatting (3DGS) \cite{kerbl20233d} employs a set of 3D points to effectively render images from given viewpoints, each characterized by a Gaussian function with 3D mean $\mathbf{\mu}_i \in \mathbb{R}^3$, covariance matrix $\Sigma_i \in \mathbb{R}^{3\times3}$, opacity value $\alpha_i \in \mathbb{R}$, RGB color values $\mathbf{c}_i \in \mathbb{R}^3$: 
\begin{equation}
    o_i(\mathbf{x}) = \alpha_i \times \exp\left( -\frac{1}{2} (\mathbf{x} - \mathbf{\mu}_i)^T \Sigma_i^{-1} (\mathbf{x} - \mathbf{\mu}_i) \right), \ \ 
    \Sigma_i = R_iS_i S^T_i R^T_i.
\label{eq:Gaussian}
\end{equation}
Given a 3D position $\mathbf{x}$, $o_i(\mathbf{x})$ represents current opacity value contributed by the $i$-th Gaussian. To facilitate optimization, $\Sigma_i$ is factorized into the product of a scaling matrix $S_i$, represented by scale factors $\mathbf{s}_i \in \mathbb{R}^3$, and a rotation matrix $R_i$ encoded by a quaternion $\mathbf{r}_i \in \mathbb{R}^4$.
% \begin{equation}
% \Sigma_i = R_iS_i S^T_i R^T_i.
% \end{equation}
3D Gaussians are then projected onto a 2D image plane according to elliptical weighted average (EWA) \cite{zwicker2002ewa} to render images for given views. Color $\mathbf{\bar{C}}(\mathbf{u})$, depth ${\bar{D}}(\mathbf{u})$, and normal $\mathbf{\bar{N}}(\mathbf{u})$ at pixel $\mathbf{u}$ is rendered by $N$ projected and ordered Gaussians using point-based $\alpha$-blending:
\begin{equation}
    \{\mathbf{\bar{C}}, \bar{D}, \mathbf{\bar{N}} \}(\mathbf{u}) = \sum_{i \in N} T_i o_i \{\mathbf{c}_i, d_i, \mathbf{n}_i \},
\label{eq:blending}
\vspace{-0.2cm}
\end{equation}
where $T_i = \prod_{j=1}^{i-1} (1 - o_j)$, depth $d_i$ is the distance between camera center and the ray-Gaussian intersection plane, and Gaussian’s normal $\mathbf{n}_i$ is approximated as the normal of the ray-Gaussian intersection plane.

\begin{figure*}[t]
\includegraphics[width=12.5cm]{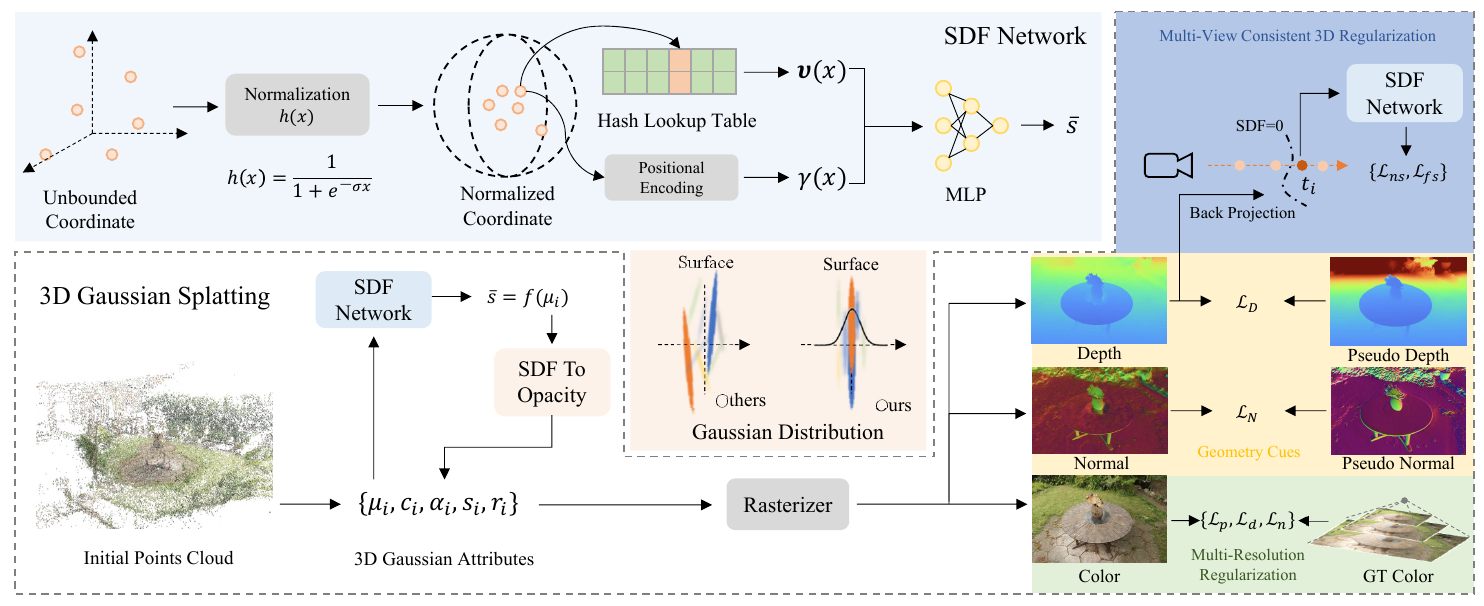}
\centering
\vspace{-0.2cm}
\caption{\textbf{Overview.} MonoGSDF synergizes SDF and Gaussian representations by converting queried SDF values into opacity, encouraging Gaussians to align with surfaces. It uses standard rasterization to render color, depth, and normals, enhanced by geometry cues and a multi-resolution strategy. SDF is jointly trained with 3D supervision from rendered depth.
}
\label{fig:pipeline}
\vspace{-0.3cm}
\end{figure*}

\paragraph{Neural Signed Distance Field.}
While 3DGS excels in novel view synthesis, its discrete nature limits surface reconstruction. In contrast, SDF provide a continuous, watertight surface as the zero-level set. We use a hash-based neural representation and jointly train the SDF with 3DGS, enabling more precise and efficient surface extraction.
To better encode the scene geometry, we choose One-blob $\gamma(\mathbf{x})$ \cite{muller2019neural} as positional embedding and a multi-resolution hash-based feature grid $\mathcal{V}=\{\mathcal{V}^l\}^L_{l=1}$ \cite{muller2022instant}. Feature $\mathcal{V}(h(\mathbf{x}))$ at any 3D point $\mathbf{x}$ are queried via trilinear interpolation. The geometry decoder $f$ is an MLP which maps the 3D coordinate to an SDF value $\bar{s}$:
\begin{equation}
    f(\mathbf{x})=f(\gamma(h(\mathbf{x})), \mathcal{V}(h(\mathbf{x}))) \mapsto \bar{s}.
\label{eq:SDF}
\end{equation}
where $h$ is a coordinate normalization. Therefore, for each 3D Gaussian located at $\mathbf{\mu}_i$, its corresponding SDF value can be written as $\bar{s}_i=f(\mathbf{\mu}_i)$.

\subsection{Gaussian-Guided Signed Distance Field}
\label{sec:optimization}
\paragraph{Connect Gaussians with SDF.}
As illustrated in Figure\ \ref{fig:pipeline}, our method begins by querying SDF values with each Gaussian 3D mean $\mathbf{\mu}_i$ using Eq.\ \ref{eq:SDF}, where zero values correspond to surface points. While 3DGSR \cite{lyu20243dgsr} introduced a bell-shaped function to map SDF value to Gaussian opacity, this approach assigns an opacity value $\alpha_i=0.25$ to surface points ($\bar{s}=0$), which contradicts the physical reality that surface points should exhibit maximum opacity ($\alpha_i=1$). To address this limitation, we propose a Gaussian-shaped transformation that maintains both mathematical simplicity and computational efficiency while ensuring physically plausible opacity values $g(\bar{s})$:
\begin{equation}
    g(\bar{s}_i) = \exp(-(\beta \times \bar{s}_i)^2) \mapsto \alpha_i,
\label{eq:CDF}
\end{equation}
where $\beta$ is a hyperparameter. 
While previous 3DGS methods \cite{kerbl20233d, huang20242d, lu2024scaffold} optimize opacities through alpha blending without considering spatial distribution constraints, these approaches often result in limited geometric accuracy. Our method enforces a better distribution of Gaussians along the surface, leading to high geometric fidelity.

\paragraph{Gaussian-Guided Normalization}
Our rendering pipeline follows the standard Gaussian rasterization process, generating color $\mathbf{\bar{C}}$, depth $\bar{D}$ and $\mathbf{\bar{N}}$ maps from Gaussian attributes, with the depth map subsequently used for SDF training. While 3DGSR \cite{lyu20243dgsr} and GSDF \cite{yu2024gsdf} have adopted similar approaches, their grid-based representations necessitate predefined scene bounding boxes for volume initialization, restricting their applicability to object-level or small-scale scene reconstruction. Although MipNeRF \cite{barron2022mip} introduced a fixed coordinate contraction method for unbounded scenes, its mapping of infinite space to [-2, 2] conflicts with our hash-based feature field's requirement for (0, 1) input. Moreover, MipNeRF's linear transformation within a limited region of interest introduces substantial distortion in peripheral areas.
To address these limitations and to enable scalable reconstruction of real-world unbounded scenes, our method introduces a Gaussian-guided normalization strategy. We first establish a bounding box based on the Gaussians, then apply a novel sigmoid-like normalization function that maps 3D point $\mathbf{x} \in \mathbb{R}^3$ to a fixed $(0, 1)$ range:
\begin{equation}
    h(\mathbf{x}) = 1 / (1 + \exp(-\sigma\mathbf{x})),
\label{eq:cor-nor}
\end{equation}
where $\sigma=2/B$, $B \in \mathbb{R}^{3}$ is the bounding box size of initial Gaussians. This approach ensures the volume containing the primary Gaussians occupies the majority of the grid space. During Gaussian densification, while new Gaussians may extend beyond the initial bounding box, they remain within the normalized scope. Our sigmoid-like normalization maintains near-linear transformation for Gaussians within the primary bounding box while effectively contracting distant Gaussians, achieving minimal distortion.

\subsection{Optimization}
\paragraph{Geometry Cues.}
Precisely reconstructing real-world scenes from only monocular images remains a challenging problem.
To further improve the reconsruction quality, we incorporate dense pseudo disparities $\hat{Z}$ from a depth estimator\footnote{We use DepthAnything V2 \cite{yang2024depth}.}. However, these suffer from scale ambiguity and view inconsistency. While DN Splatter \cite{turkulainen2024dn} aligns depth via $\bar{D} = s\hat{D} + t$, it overlooks disparity shifts, causing depth distortion. We address this by refining pseudo depth as $\hat{D} = a / (s\hat{Z} + t) + b$ using rendered depth, then derive pseudo normals $\hat{\mathbf{N}} = \nabla\hat{D}/|\nabla\hat{D}|$ for stable Gaussian supervision.
\begin{equation}
    \mathcal{L}_{D} = \sum |({a/(s \times \hat{Z}+t)+b-\bar{D}}|, \ \ 
    \mathcal{L}_{N} = \sum (1 - \mathbf{\hat{N}} \cdot \mathbf{\bar{N}}).
\label{eq:depthany}
\end{equation}
where $s, t$ and $a, b$ represent learnable scale and shift parameters, and $\mathbf{\bar{N}}$ is rendered Gaussian normal. Therefore, the overall geometry regularization is $\mathcal{L}_{geo} = \lambda_{D}\mathcal{L}_{D} + \lambda_{N}\mathcal{L}_{N}$. 

% \begin{equation}
%     \mathcal{L}_{N} = \sum (1 - \mathbf{\hat{N}} \cdot \mathbf{\bar{N}}).
% \end{equation}
% where $\mathbf{\bar{N}}$ is rendered Gaussian normal.
% To achieve high-quality planar reconstruction, we implement a Total Variation (TV) regularization \cite{Niemeyer2021Regnerf} that operates directly in 3D space. Unlike DN Splatter \cite{turkulainen2024dn} which applies TV regularization on 2D depth maps, our approach first projects depth values into world coordinates using camera intrinsic $K$ and extrinsic $T$ to generate a 3D point map $\mathbf{\bar{P}} = T^{-1}K^{-1}\bar{D}$, then computes the TV loss in 3D space:
% \begin{equation}
%     \mathcal{L}_{tv} = \sum_{i,j} g_{c} \cdot(|\mathbf{\bar{P}}_{i+1,j}-\mathbf{\bar{P}}_{i,j}| + |\mathbf{\bar{P}}_{i,j+1}-\mathbf{\bar{P}}_{i,j}|),
% \end{equation}
% where $i, j$ are the indices of pixels, $g_{c}=\exp(- \nabla \mathbf{C})$ is the gradient of color image. 
% This 3D-aware regularization ensures planar structures remain geometrically flat in world space, rather than merely producing smooth depth values in image space.

While the geometry cues offer valuable 2D supervision for improving depth and normal, they fail to constrain the 3D Gaussian distribution effectively. We address this limitation through the SDF, which enforces tied surface alignment of Gaussians.

\paragraph{Multi-View Consistent 3D Regularization.}
To train this hybrid model, enforcing all Gaussians to satisfy $\bar{s}_i=0$ (leading to $\alpha_i=1$) degrades rendering quality, as high-quality renderings require opacity variation, and sparse SDF supervision at Gaussian positions is insufficient. Dense supervision with monocular cues like MonoSDF \cite{yu2022monosdf} suffers from scale ambiguity and lacks multi-view consistency, while methods like 3DGSR \cite{lyu20243dgsr} and GSDF \cite{yu2024gsdf} only offer 2D supervision through rendered Gaussian depth maps. Our approach leverages monocular cues and the 3D Gaussians' advantages, where the Gaussian-rendered depth $\bar{D}$ ensures multi-view consistency and provides reliable 3D SDF supervision via back-projection.

More specifically, we first sample $M$ pixels $\{\mathbf{u} = \mathbf{u}_i\ |\ \mathbf{u}_i \in \mathbb{R}^2, {\text{for}\ i=1, 2, ...,M\}}$ from the rendered Gaussian depth map $\bar{D}$. For each ray casts from a pixel, we sample $K=K_n + K_f$ points $\{t_k\}$ from the camera center to the surface, where $K_n$ means near surface samples and $K_f$ means free space samples. SDF values of each point can be queried as $\{\bar{s}_k\}$.
For near surface samples within the truncation region $tr$, i.e. $S_{tr}=\{|\bar{D}(\mathbf{u}) - t_k| \le tr\}$, we use the distance between the sampled point  $t_k$ and its surface $\bar{D}(\mathbf{u}_j)$ as an approximation of ground truth SDF value for supervision, and for points that are far from the surface $S_{fs}=\{\bar{D}(\mathbf{u}) - t_k > tr\}$, we apply a free-space supervision:
\begin{equation}
\vspace{-0.2cm}
    \mathcal{L}_{ns} = \sum_{\mathbf{u}_j \in \mathbf{u}} \sum_{t_k \in S_{tr}} ||\bar{s}_k - (\bar{D}(\mathbf{u}_j) - t_k)||_2, \ \ 
    \mathcal{L}_{fs} = \sum_{\mathbf{u}_j \in \mathbf{u}} \sum_{t_k \in S_{fs}} ||\bar{s}_k - 1||_2.
% \vspace{-0.2cm}
\end{equation}

% \begin{equation}
%     \mathcal{L}_{fs} = \sum_{\mathbf{u}_j \in \mathbf{u}} \sum_{t_k \in S_{fs}} ||\bar{s}_k - 1||_2.
% \end{equation}
Therefore, the overall SDF regularization is $\mathcal{L}_{sdf} = \lambda_{ns}\mathcal{L}_{ns}+\lambda_{fs}\mathcal{L}_{fs}$.

\paragraph{Multi-Resolution Regularization.}
Unlike vanilla 3DGS \cite{kerbl20233d} with L1/SSIM losses and opacity-based pruning, our method links opacity to spatial position, eliminating \emph{Opacity Resets} but risking Gaussian redundancy. Instead of frequency-based strategies \cite{zhang2024fregs} that may blur structure, we introduce wavelet-based multi-resolution supervision: $\mathbf{C}_l = \mathcal{W}(\mathbf{C}, l)$, progressively refining detail while preserving structure. $\mathbf{C}$ is ground truth color images, $\mathcal{W}$ represents wavelet transform and $l$ denotes the level of transformation. This stabilizes training and enables coarse-to-fine Gaussian distribution, preventing redundant Gaussians. The photometric loss is $\mathcal{L}_p = 0.8|\mathbf{C}_l - \bar{\mathbf{C}}| + 0.2 \cdot SSIM(\mathbf{C}_l, \bar{\mathbf{C}})$.

\paragraph{Objective Function.}
During training, we jointly optimize explicit Gaussians and implicit SDF. We apply distortion $\mathcal{L}_{d}$ and depth-normal $\mathcal{L}_{n}$ regularization as defined in GOF \cite{yu2024gaussian}. The overall objective function is defined as:
$\mathcal{L} = \mathcal{L}_{p} + \mathcal{L}_{geo} + \mathcal{L}_{sdf} + \lambda_d \mathcal{L}_{d} + \lambda_n \mathcal{L}_{n}$.

\paragraph{Pruning.}
In our framework, opacity is tied to spatial position via SDF. We adopt a geometry-aware pruning strategy that removes Gaussians with $\bar{s}_i > tr$, replacing conventional opacity-based pruning with a surface-aware, physically grounded criterion.

% move to supp
% \paragraph{Mesh Extraction.}
% Previous Marching Cubes-based approaches \cite{yu2024gsdf, lyu20243dgsr} must densely search both occupied and free space, leading to inefficiencies and resolution limitations, especially in unbounded scenes with coordinate distortions. In contrast, our method adapts Marching Tetrahedral \cite{kulhanek2023tetra} and exploits 3DGS’s spatial cues to constrain the search to regions only near Gaussians, avoiding unnecessary exploration of empty space while achieving superior surface alignment and reconstruction efficiency.

% \begin{equation}
%     \mathcal{L}_{c} = \sum_{i,j} (1 - \lambda)||\mathbf{I} - \mathbf{\bar{I}}||_1 + \lambda SSIM(\mathbf{I} - \mathbf{\bar{I}})
% \end{equation}
% \begin{equation}
%     \mathcal{L}_{\text{dd}} = \sum_{i,j} \omega_i\omega_j||t_i - t_j||_1
% \end{equation}
% where $i,j$ index over Gaussians contributed to the ray and $t_i$ is the depth of the intersection point with Gaussians.
% \begin{equation}
%     \mathcal{L}_{\text{n}} = \sum_{i,j} ||1 - \mathbf{\bar{N}}^{T}\mathbf{\hat{N}}||_1
% \end{equation}
% where $\mathbf{\bar{ N}}$ is rendered normal map, and $\mathbf{\hat{N}}$ is the normal estimated by the gradient of the depth map.
\input{tables/dtu}

\begin{figure*}[t]
\includegraphics[width=12cm]{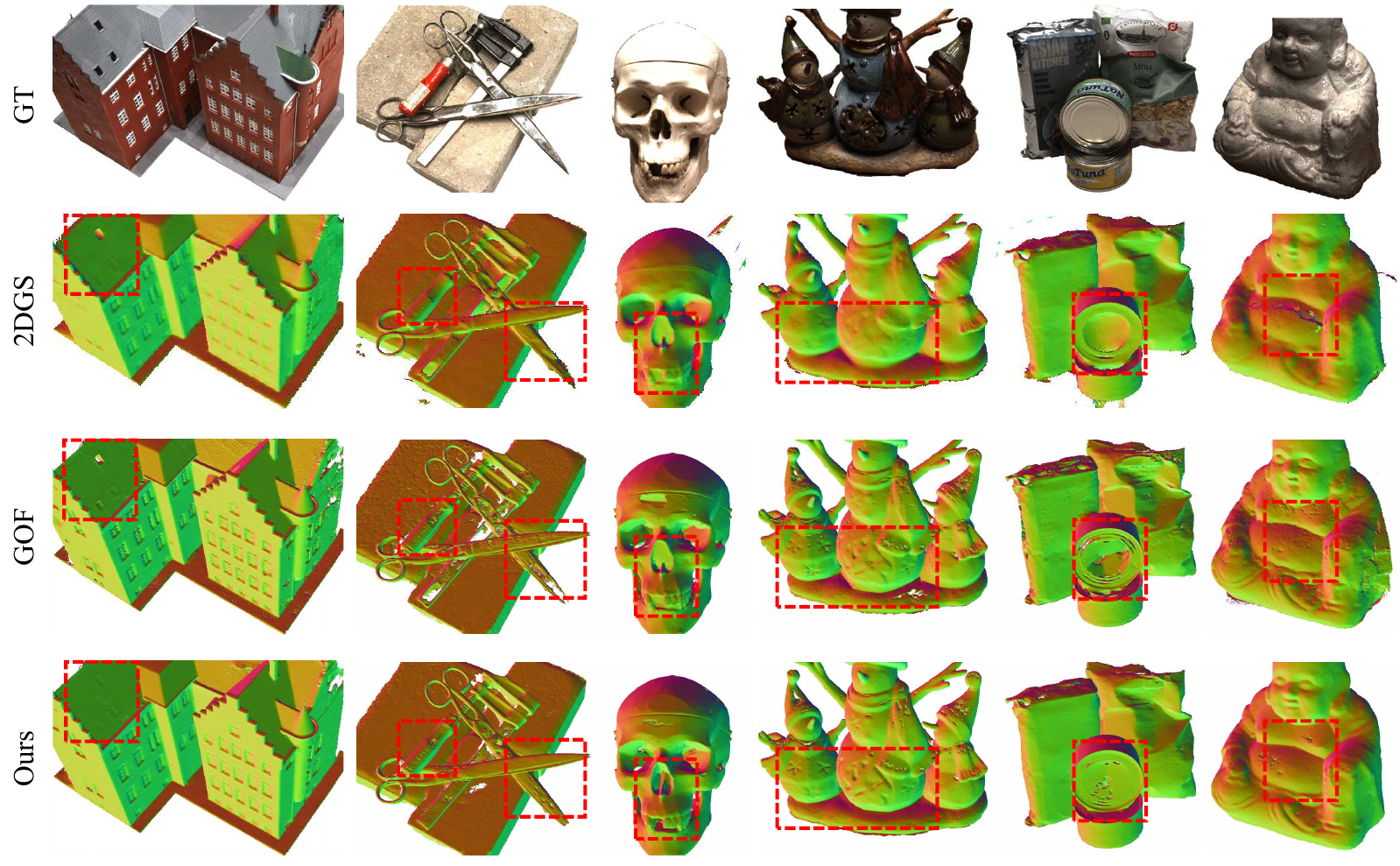}
\centering
\vspace{-0.2cm}
\caption{\textbf{Surface Reconstruction on the DTU~\cite{jensen2014large}.} We show normal maps from the reconstructed meshes.} 
\label{fig:dtu}
\vspace{-0.2cm}
\end{figure*}

%% file: tables/dtu.tex
\setlength\tabcolsep{0.5em}
\begin{table}[t]
\centering

% \vspace{-0.2cm}
\resizebox{\textwidth}{!}{
\begin{tabular}{@{}lcccccccccccccccclc}
\hline
 \multicolumn{2}{c}{} & 24 & 37 & 40 & 55 & 63 & 65 & 69 & 83 & 97 & 105 & 106 & 110 & 114 & 118 & 122 & & Mean \\ 
 \cline{1-19} 
 NeRF~\cite{mildenhall2021nerf} & & 1.90 & 1.60 & 1.85 & 0.58 & 2.28 & 1.27 & 1.47 & 1.67 & 2.05 & 1.07 & 0.88 & 2.53 & 1.06 & 1.15 & 0.96 & & 1.49\\
 VolSDF~\cite{yariv2021volume} & &  1.14 &  1.26 &  0.81 & 0.49 & 1.25 &  \tbest0.70 & \tbest 0.72 & \tbest 1.29 & \tbest 1.18 &  \tbest 0.70 & 0.66 & 1.08 &  0.42 & \tbest 0.61 &  0.55 & & 0.86\\
 NeuS~\cite{wang2021neus} & &  1.00 & 1.37 & 0.93 &  0.43 & 1.10 &  \sbest 0.65 &   \best 0.57 &  1.48 &  \sbest 1.09 &  0.83 &  \tbest 0.52 &  1.20 & \best 0.35 &  \best 0.49 &  0.54 & &  0.84\\
 N-angelo~\cite{li2023neuralangelo} & & \tbest 0.49 & 1.05 & 0.95 & \tbest 0.38 & 1.22 & 1.10 & 2.16 & 1.68 & 1.78 & 0.93 & \best 0.44 & 1.46 & 0.41 & 1.13 & 0.97 & & 1.07\\ 
 \cline{1-19}

 SuGaR~\cite{guedon2024sugar} & & 1.47 & 1.33 & 1.13 & 0.61 & 2.25 & 1.71 & 1.15 & 1.63 & 1.62 & 1.07 & 0.79 & 2.45 & 0.98 & 0.88 & 0.79 & & 1.33\\
 2DGS~\cite{huang20242d} &&  \sbest 0.48 & 0.91 & \tbest 0.39 & 0.39 &  \tbest 1.01 &  0.83 &  0.81 &  1.36 &  1.27 &  0.76  &  0.70 & 1.40 & 0.40 &   0.76 & 0.52 && 0.80 \\
 GSurfel~\cite{dai2024high} & & 0.66 & 0.93 & 0.52 & 0.41 & 1.06 &  1.14 & 0.85 & \tbest 1.29 & 1.53 & 0.79 & 0.82 & 1.58 & 0.45 & 0.66 & 0.53 && 0.88 \\
 3DGSR~\cite{lyu20243dgsr} & & 0.68 & 0.84 & 0.70 & 0.39 & 1.16 &  0.87 & 0.77 & 1.48 & 1.26 & 0.87 & 0.69 & \best0.80 & 0.42 & 0.64 & 0.60 && 0.81 \\ 
 GS-Pull~\cite{zhang2025neural} & & 0.51 & \best 0.56 & 0.46 & 0.39 & \best 0.82 & \best 0.67 & 0.85 & 1.37 & 1.25 & 0.73 & 0.54 & 1.39 & \best 0.35 & 0.88 & \best0.42 && \tbest0.75 \\ 
 GOF~\cite{yu2024gaussian} & & 0.50 & \tbest 0.82 & \sbest 0.37 & \sbest 0.37 & 1.12 &  0.74 & 0.73 & \best 1.18 & 1.29 & \sbest 0.68 & 0.77 & \tbest 0.90 & 0.42 & 0.66 & \tbest 0.49 && \sbest 0.74 \\
 Ours & & \best 0.45 & \sbest 0.65 & \best 0.36 & \best0.36 & \sbest0.94 &  \tbest0.70 & \sbest0.67 & \sbest1.27 & \best0.99 & \best0.63 & \sbest0.49 & \sbest0.84 & \tbest0.39 & \sbest0.53 & \sbest 0.47 && \best 0.65 \\
 \hline
\end{tabular}
}
\vspace{+0.2cm}
\caption{\textbf{Quantitative Comparison on the DTU~\cite{jensen2014large}}. We show the Chamfer distance. For N-angelo \cite{li2023neuralangelo}, we report the results from UniSDF \cite{wang2023unisdf} reproduction and we show the vanilla results in supp. mat.. }
\label{tab:dtu_result}
% \vspace{-0.3cm}
\end{table}

%% file: sec/4_experiments.tex
\section{Experiments}
\label{sec:experiments}

\subsection{Experimental Settings} 
\paragraph{Datasets.} We comprehensively evaluate our method on three public datasets: \emph{DTU} dataset \cite{jensen2014large} which consists of indoor object scans; \emph{Tanks and Temples} \cite{knapitsch2017tanks} which features six unbounded outdoor scenes; and \emph{Mip-NeRF} 360 \cite{barron2022mip} for measuring novel view synthesis quality.

\paragraph{Baselines.} We compare our proposal against several state-of-the-art methods. Among implicit methods, we compare against NeRF \cite{mildenhall2021nerf}, VolSDF \cite{yariv2021volume}, NeuS \cite{wang2021neus}, N-angelo \cite{li2023neuralangelo}, GeoNeus \cite{fu2022geo}, Instant NGP \cite{mueller2022instant} and MipNeRF 360 \cite{barron2022mip}. As for explicit Gaussian methods, we compare against 3DGS \cite{kerbl20233d}, SuGaR \cite{guedon2024sugar}, Mip-Splatting \cite{yu2024mip}, 2DGS \cite{huang20242d}, GSurfel \cite{dai2024high}, 3DGSR \cite{lyu20243dgsr}, GS-Pull \cite{zhang2025neural} and GOF \cite{yu2024gaussian}.

\paragraph{Metrics.} We follow common practice and report surface accuracy as Chamfer Distance and F1-score on DTU and Tanks and Temples, respectively. We measure the visual fidelity of the synthesized novel views with PSNR, SSIM and LPIPS~\cite{zhang2018unreasonable} on Mip-NeRF 360.

\paragraph{Implementation.} We perform single GPU training (NVIDIA 3090) and use by default $\lambda_{d}=100, \lambda_{n}=0.05, \lambda_{ns}=1000, \lambda_{fs}=10, \lambda_{depth}=0.05, \lambda_{normal}=0.1$ and $\beta=100$. We train 30000 iterations like other methods and do not require extra training for SDF. For more implementation details, please refer to supp.\ mat..

\input{tables/tnt}

\subsection{Geometry Evaluation}
\begin{figure*}[t]
\includegraphics[width=12cm]{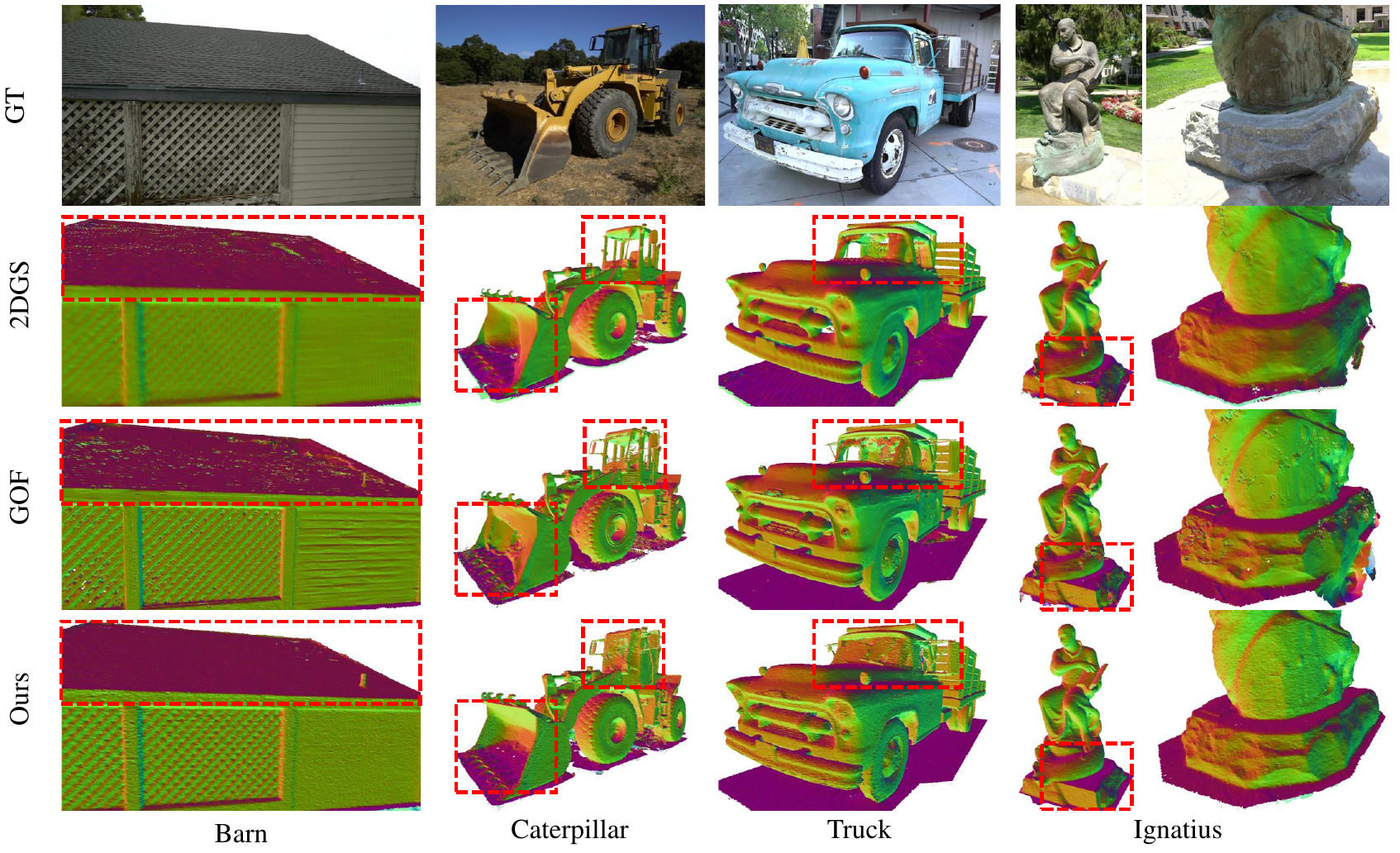}
\centering
% \vspace{-0.2cm}
\caption{\textbf{Surface Reconstruction on the Tanks and Temples~\cite{knapitsch2017tanks}.} We show rendered normal maps from reconstructed meshes.} 
\label{fig:tnt}
% \vspace{-0.2cm}
\end{figure*}

We evaluate our method on the DTU dataset \cite{jensen2014large}, achieving the lowest Chamfer Distance and outperforming all 3DGS-based and implicit methods in quality (Table \ref{tab:dtu_result}). For N-angelo \cite{li2023neuralangelo}, we report results from \cite{wang2023unisdf} due to unverified original values. As shown in Figure \ref{fig:dtu}, our method yields smoother, more complete surfaces—especially in sparse and flat regions—compared to noisy outputs from 2DGS \cite{huang20242d} and GOF \cite{yu2024gaussian}.
On the Tanks and Temples dataset \cite{knapitsch2017tanks}, as shown in Table~\ref{tab:tnt}, our method matches the performance of leading implicit approaches \cite{li2023neuralangelo} while cutting training time from 12+ to around 3 hours. We also slightly outperform GOF \cite{yu2024gaussian}, with faster mesh extraction and similar training speed (Table \ref{tab:runtime}). Figure \ref{fig:tnt} shows our superior handling of flat and transparent surfaces. Additional results on Mip-NeRF 360 \cite{barron2022mip}, DTU \cite{jensen2014large}, and Tanks and Temples \cite{knapitsch2017tanks} are in the supp. mat..

% Similarly, on the Mip-NeRF 360 dataset \cite{barron2022mip}, Figure \ref{fig:360} shows our method reconstructs detailed meshes, while 2DGS produces overly smooth meshes and GOF generates noisy artifacts on flat surfaces.

\input{tables/mipnerf-360}

\subsection{Novel View Synthesis}
We further compare our method with state-of-the-art novel view synthesis (NVS) techniques on the Mip-NeRF 360 dataset \cite{barron2022mip}. The quantitative results presented in Table \ref{tab:mipnerf360} demonstrate that our approach achieves remarkable rendering performance. Our method effectively aligns the Gaussians with the surface, leading to better surface reconstruction without compromising rendering quality.

\input{tables/ablation}
\subsection{Runtime and Ablation Study}

\begin{figure}[t]
\includegraphics[width=9cm]{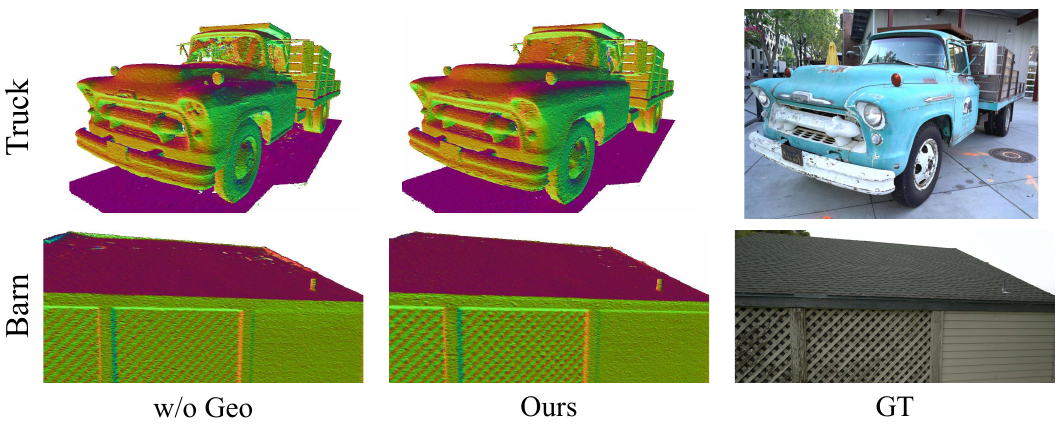}
\centering
\vspace{-0.2cm}
\caption{\textbf{Ablation Study on Geometry Regularization.} We show the reconstructed meshes with and without the geometry regularization.} 
\label{fig:ablation}
\vspace{-0.2cm}
\end{figure}

\paragraph{Runtime Analysis.}
Table \ref{tab:runtime} reports the runtime analysis on the Barn scene. Our method integrates seamlessly with different Gaussian rasterizers. “Ours (2D)” refers our method to using the 2DGS rasterizer. While SDF guidance enhances surface reconstruction, our approach achieves competitive training times and notably faster mesh extraction than baselines such as 2DGS and GOF, since our SDF network directly outputs values for any query point, whereas GOF relies on ray tracing and 2DGS requires TSDF fusion.

% \subsubsection{Ablations on Different Regularizations}
\paragraph{Ablations on Geometry Regularization.} 
Table \ref{tab:ablation} presents ablation studies on the impact of different geometry regularization terms, showing similar performance gaps for pseudo normal and depth maps. Evaluating GOF with geometry cues only ("GOF+Geo") shows modest improvements (0.52). To highlight that geometry cues alone aren't the main factor, we increase $\lambda_{D}$ and $\lambda_{N}$ to 0.5, leading to performance degradation due to their adverse effects. Our complete pipeline achieves an 8\% improvement, demonstrating the synergistic effect of our approach. Figure \ref{fig:ablation} illustrates the effectiveness of our geometry regularization, especially in improving surface consistency and reconstruction quality in transparent and flat regions.

\paragraph{Ablations on Normalization Function.} 
MipNeRF 360’s non-linear normalization \cite{barron2022mip} introduces distortions and conflicts with tinycudann’s $(0,1)$ input requirement. In contrast, our near-linear normalization within the bounding box compresses outliers with minimal distortion. For fairness, we scale MipNeRF’s normalization output to $(0,1)$ and compare with ours in Table \ref{tab:ablation}, where our method ("Ours") achieves superior performance.

\paragraph{Ablations on SDF to Opacity Function.} 
We compare 3DGSR's SDF-to-Opacity function \cite{lyu20243dgsr} ("w/ 3DGSR SDF2O") with our method in Table \ref{tab:ablation}. Our approach achieves a higher F1-score, providing a simpler, more efficient solution with full opacity, while 3DGSR's opacity is capped at 0.25, limiting its effectiveness in unbounded outdoor scenes.

\paragraph{Ablations on SDF Fields.} 
Table \ref{tab:ablation} shows that our Gaussian-based implicit surface reconstruction outperforms GOF, and when integrated with the 2DGS rasterizer (“Ours (2D)”), it consistently achieves higher F1-scores across settings.

%% file: tables/tnt.tex
\begin{table}[t]
\centering

% \vspace{-0.2cm}
\resizebox{0.9\columnwidth}{!}{
\begin{tabular}{@{}l|ccc|cccc}
 \cline{1-8}
 & \multicolumn{3}{c@{}|}{Implicit} & \multicolumn{4}{c@{}}{Explicit} \\ 
 & NeuS \cite{wang2021neus} & GeoNeus \cite{fu2022geo} & N-angelo \cite{li2023neuralangelo} & SuGaR \cite{guedon2024sugar} & 2DGS \cite{huang20242d} & GOF \cite{yu2024gaussian} & Ours\\ 
 \hline
Barn & 0.29 &  0.33 &  \best 0.70  & 0.14 & 0.36 & \tbest 0.51 & \sbest 0.56\\
Caterpillar & 0.29 & 0.26 &  \tbest 0.36 & 0.16 & 0.23 & \best 0.41 & \sbest0.38\\
Courthouse & 0.17 & 0.12 &  \sbest 0.28 & 0.08 & 0.13 & \sbest 0.28 & \best0.29\\
Ignatius &   \sbest 0.83 & \tbest 0.72 &  \best 0.89 & 0.33 & 0.44 & 0.68 & \tbest 0.72\\
Meetingroom & 0.24 & 0.20 &  \best 0.32 &  0.15 & 0.16 & \sbest0.28 & \tbest 0.25\\
Truck &  0.45 &  0.45 &  \tbest 0.48 &  0.26 & 0.26 & \sbest 0.59 & \best 0.62\\ 
\hline
Mean & 0.38 & 0.35 &  \best 0.50 & 0.19 & 0.30 & \tbest0.46 & \sbest 0.47\\
\cline{1-8}
\end{tabular}
}
\vspace{+0.2cm}
\caption{\textbf{Quantitative Results on the Tanks and Temples~\cite{knapitsch2017tanks}}.We show the F1-score.}
\label{tab:tnt}
% \vspace{-0.3cm}
\end{table}

%% file: tables/mipnerf-360.tex
\begin{table}[t]
\centering
\resizebox{\columnwidth}{!}{
\begin{tabular}{@{}l|cccccccccc@{}}
\toprule
Metric & NeRF~\cite{mildenhall2021nerf} & InstantNGP~\cite{mueller2022instant} & MipNeRF360~\cite{barron2022mip} & 3DGS~\cite{kerbl20233d} & SuGaR~\cite{guedon2024sugar} & Mip-Splat~\cite{yu2024mip} & 2DGS~\cite{huang20242d} & GS-Pull~\cite{zhang2025neural} & GOF~\cite{yu2024gaussian} & Ours \\
\midrule
PSNR~$\uparrow$ & 24.15 & 26.03 & \best 28.10 & 27.53 & 26.18 & 27.78 & 27.16 & 27.27 & \sbest 27.81 & \sbest 27.81 \\
SSIM~$\uparrow$ & 0.624 & 0.723 & 0.804 & \tbest 0.826 & 0.768 & 0.825 & 0.811 & 0.814 & \best 0.837 & \sbest 0.836 \\
LPIPS~$\downarrow$ & 0.443 & 0.294 & 0.232 & \tbest0.212 & 0.291 & 0.220 & 0.244 & 0.230 & \sbest 0.193 & \best 0.189 \\
\bottomrule
\end{tabular}
}
\vspace{+0.2cm}
\caption{\textbf{Quantitative Results on Mip-NeRF 360~\cite{barron2022mip}.} We evaluate the quality of rendered images against ground-truth images using PSNR, SSIM, and LPIPS.
}
\label{tab:mipnerf360}
% \vspace{-0.2cm}
\end{table}

%% file: tables/ablation.tex
\begin{table}[ht]
\centering

\vspace{-0.1cm}
\resizebox{0.5\columnwidth}{!}{
\begin{tabular}{@{}l|ccccc}
 \hline
 \textbf{Run Time} & 2DGS & Ours (2D) & GOF & Ours & \\ 
 \hline
 Training & 17 min & 39 min & 2.8 h & 3.1 h & \\
 \hline
 Meshing & $\sim$20 min & 5 min & $\sim$30 min & 5 min &\\
 \hline

\end{tabular}
}
\vspace{+0.1cm}
\caption{\textbf{Runtime.} We report the training and meshing time. Compared to GOF~\cite{yu2024gaussian}, we avoid ray tracing for SDF evaluation, enabling fast meshing.}
\label{tab:runtime}
% \vspace{-0.2cm}
\end{table}

\begin{table}[ht]
\centering

% \vspace{-0.2cm}
\resizebox{0.8\columnwidth}{!}{
\begin{tabular}{@{}l|cc|ccc|c}
 \hline
 \textbf{Method} & 2DGS & Ours (2D) & $\lambda_{D}=0$ & $\lambda_{N}=0$ & $\lambda_{D}=0$, $\lambda_{N}=0$ & w/ MipNeRF NF\\ 
 \hline
 F1 Score & 0.36 & 0.40 & 0.52 & 0.51 & 0.50 & 0.52 \\
 \hline
 \textbf{Method} & GOF & GOF+Geo & $\lambda_{D}=0.5$ & $\lambda_{N}=0.5$ & \textbf{Ours} & w/ 3DGSR SDF2O\\ 
 \hline
 F1 Score & 0.51 & 0.52 & 0.45 & 0.47 & \textbf{0.56} & 0.24 \\
 \hline

\end{tabular}
}
\vspace{+0.1cm}
\caption{\textbf{Quantitative Ablation Study.} We report Barn scene F1-scores with and without different regularization.}
\label{tab:ablation}
% \vspace{-0.2cm}
\end{table}

%% file: sec/5_conclusion.tex
\section{Conclusion}
\label{sec:conclusion}
We propose MonoGSDF, a Gaussian-based implicit surface reconstruction framework that bridges implicit and explicit representations via an SDF-to-opacity mapping. Gaussian-guided normalization stabilizes optimization in unbounded scenes, and multi-resolution training with geometric cues enables progressive and accurate reconstruction. Our method achieves SOTA performance, especially on flat and transparent surfaces where prior methods fail.

\newpage

%% file: sec/X_suppl.tex
\clearpage
\setcounter{page}{1}
% \bmvcreviewcopy{690}
% \maketitlesupplementary

\appendix
\section{Implementation Details}
\subsection{Hyperparameters}
In the following, we report implementation details and hyperparameters used for our method.

\noindent \textbf{Gaussian Splatting.} 
For hyperparameters used in the Gaussian rasterization, we follow previsous works \cite{yu2024gaussian, huang20242d, kerbl20233d}. We train our Gaussian model and SDF network jointly with 30000 iterations. We set the initial learning rate for Gaussians' position as 0.00016 and the final initial learning rate as 0.0000016. And we set learning rates for scales and rotation as 0.005, 0.001, respectively. We start the Gaussian densification after 500 iterations and until 15000 iterations. We densify Gaussians every 100 iterations and the gradient threshold for densification is 0.0002. We start the distortion loss $\mathcal{L}_{d}$ and the depth-normal loss $\mathcal{L}_{n}$ after 3000 iterations and 7000 iterations, respectively. And we set the wavelet level as 3 and gradually increase the resolution until 10000 iteration. After that, we use the full resolution for training.

\noindent \textbf{SDF Network.} 
We use two-layer MLPs and the hidden dimension is 32. We initialize $\beta$ from Eq.\ \ref{eq:CDF} as 100. We start train our SDF from 5000 iterations and until 30000 iterations. The learning rate for our SDF network is 0.002. For each iteration, we sample $M=10000$ pixels and $K_n=11, K_f=64$ points. 

\noindent \textbf{Geometry Regularization.}
We set the learning rate for $s,t, a, b$ in Eq.\ \ref{eq:depthany} as 0.01. And for Tanks and Temples, we set $\lambda_{D}=0.05, \lambda_{N}=0.1$, for DTU and Mip-NeRF 360, we set $\lambda_{D}=0.01, \lambda_{N}=0.01$.

\section{Distortion and Depth-Normal Loss}
We apply a distortion loss~\cite{huang20242d} and depth-normal loss~\cite{yu2024gaussian} as discussed in Section \ref{sec:optimization}.
The distortion loss concentrates the weight distribution along the rays by minimizing the distance between the ray-splat intersections:
\begin{equation}
    \mathcal{L}_{d} = \sum_{i,j}\omega_i\omega_j|z_i-z_j|,
\end{equation}
where $\omega_i = \,o_i(x)\prod_{j=1}^{i-1} (1 - \,o_j(x))$ is the blending weight of the $i-$th intersection and $z_i$ is the depth of the intersection points. The depth-normal loss is defined as:
\begin{equation}
    \mathcal{L}_{n} = \sum(1-\bar{\mathbf{N}}\cdot\nabla(\bar{D})),
\end{equation}
where $\nabla(\bar{D})$ is the gradient of rendered Gaussian Depth.

\section{Mesh Extraction.}
Previous Marching Cubes-based methods \cite{yu2024gsdf, lyu20243dgsr} require dense querying across both occupied and free space, resulting in inefficiencies and limited resolution—especially in unbounded scenes with coordinate distortions. In contrast, we adopt Marching Tetrahedra \cite{kulhanek2023tetra} and leverage spatial cues from 3D Gaussians to restrict the search to regions near the surface, significantly reducing unnecessary computation while improving surface alignment and reconstruction efficiency. Unlike GOF \cite{yu2024gaussian}, which performs costly ray tracing to query opacity values for each point, our method directly outputs SDF values, accelerating meshing from 30 minutes to just 5 minutes.

\section{Additional Experimental Results}
\subsection{Results on DTU and Mip-NeRF 360}
In Table \ref{tab:supp_dtu}, we present the results reported in the Neuralangelo publication \cite{li2023neuralangelo} on the DTU dataset. It is important to note that the quantitative results from the Neuralangelo paper have been flagged by other studies and discussions on the official GitHub repository as non-reproducible\footnote{Refer to this \href{https://github.com/NVlabs/neuralangelo/issues/65}{GitHub issue} for details.}. 

\begin{figure*}[t]
\includegraphics[width=11cm]{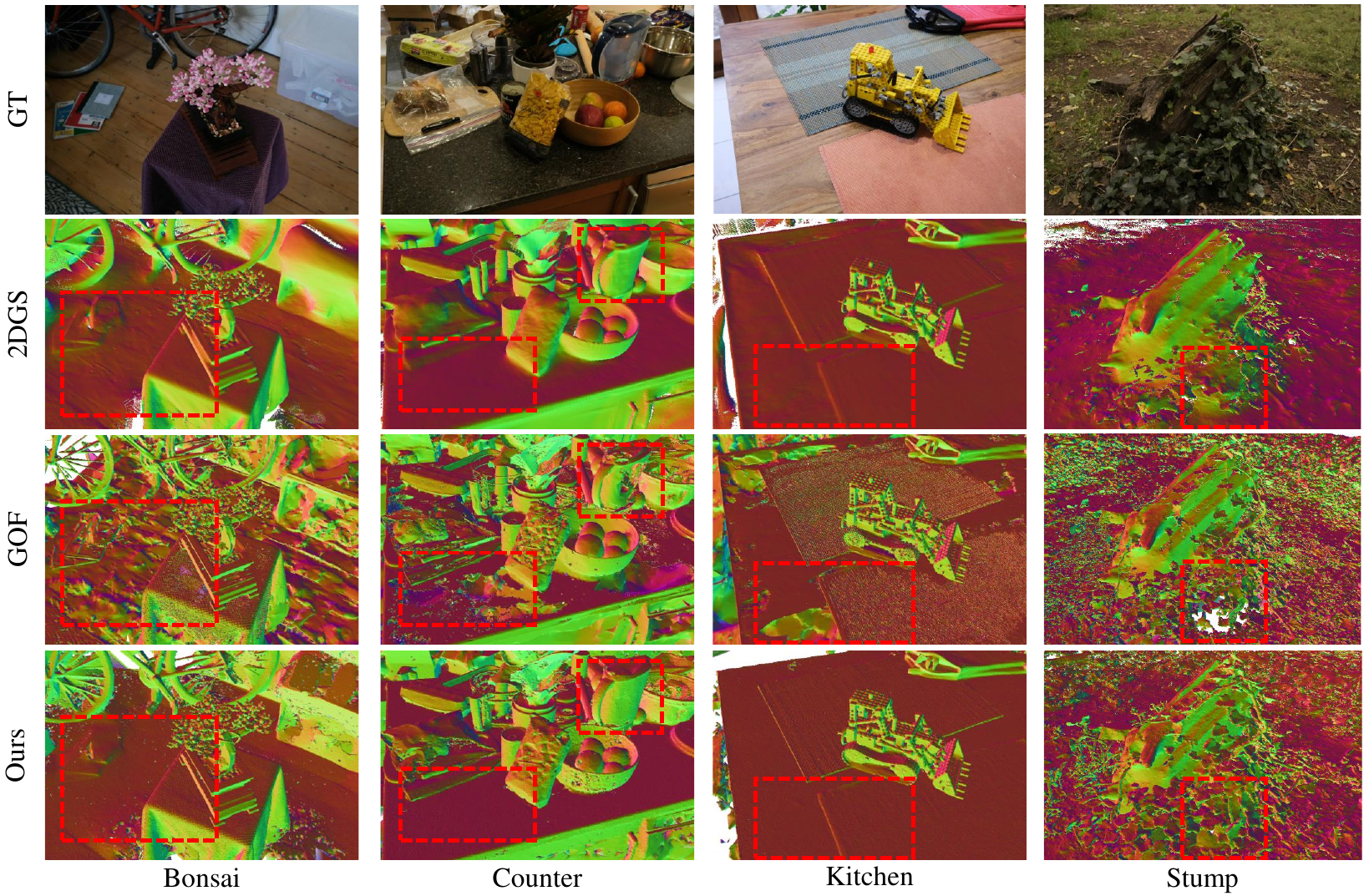}
\centering
\vspace{-0.2cm}
\caption{\textbf{Surface Reconstruction on the Mip-NeRF 360 Dataset \cite{barron2022mip}.} We show the rendered normal maps from reconstructed meshes.} 
\label{fig:360}
\vspace{-0.2cm}
\end{figure*}

\begin{figure*}[t]
\includegraphics[width=11.5cm]{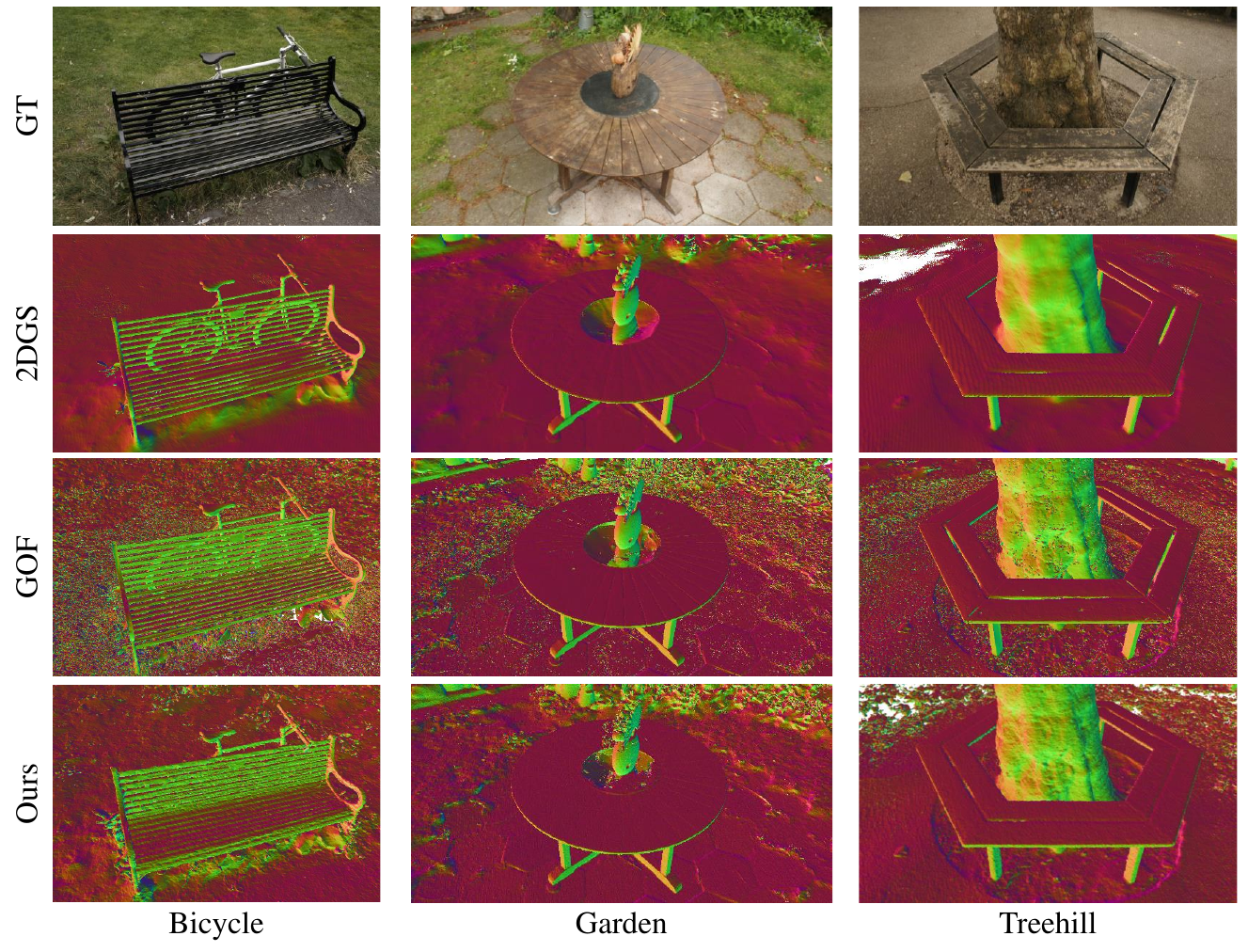}
\centering
\caption{\textbf{Additional Results on Mip-NeRF 360 \cite{barron2022mip}.} } 
\label{fig:supp_360}
% \vspace{-0.5cm}
\end{figure*}

\begin{figure*}[t]
\includegraphics[width=11.5cm]{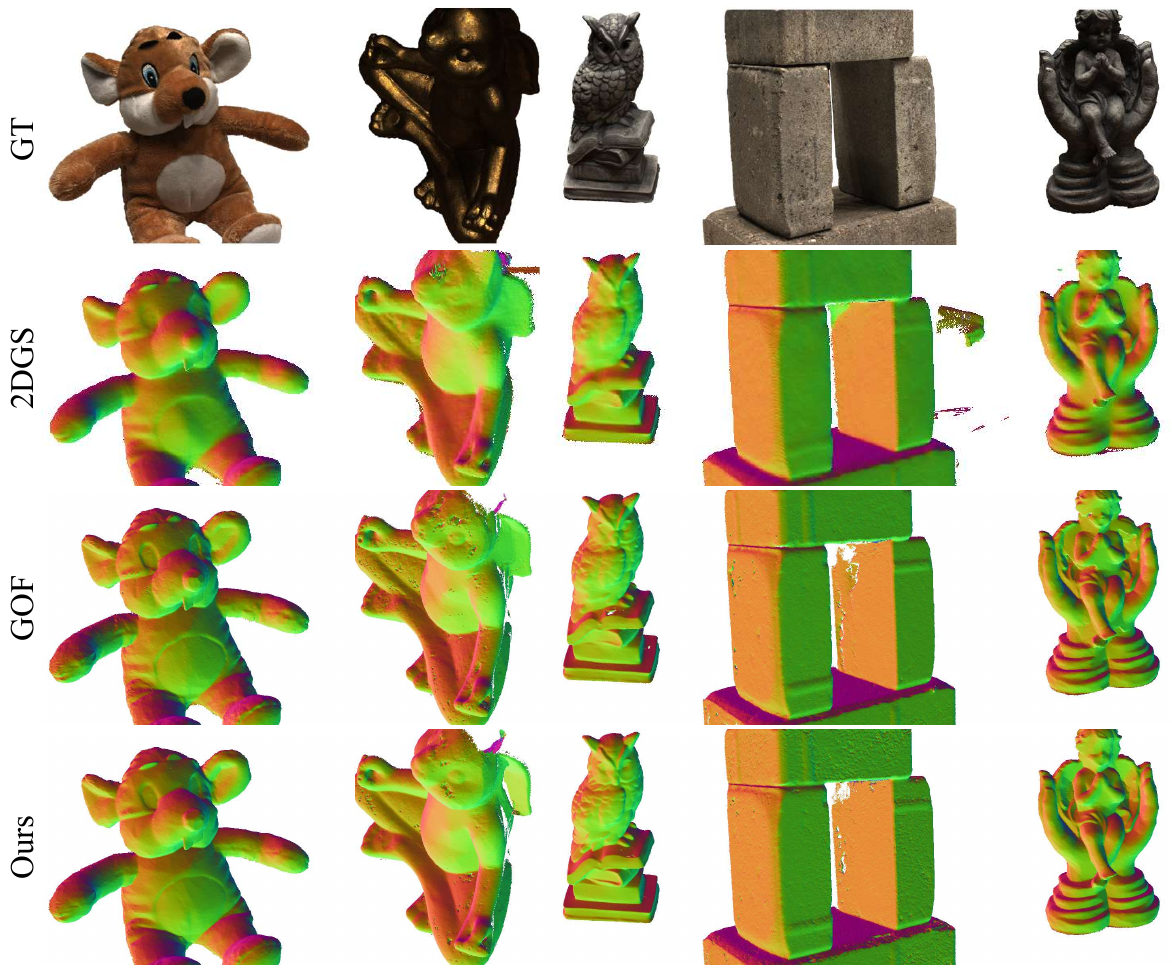}
\centering
\caption{\textbf{Additional Results on DTU \cite{jensen2014large}.} } 
\label{fig:supp_dtu}
% \vspace{-0.5cm}
\end{figure*}

To provide a more comprehensive evaluation, we include additional qualitative comparisons on the DTU dataset \cite{jensen2014large} in Figure \ref{fig:supp_dtu} and further demonstrate results on the Mip-NeRF 360 dataset \cite{barron2022mip} in Figure \ref{fig:360} and \ref{fig:supp_360}. Our method excels in handling reflective surfaces and produces reconstructions that are not only smoother but also exhibit finer details, highlighting its capability for more accurate and visually appealing scene representations.

\section{More Ablation}
\begin{figure*}[t]
\includegraphics[width=11.5cm]{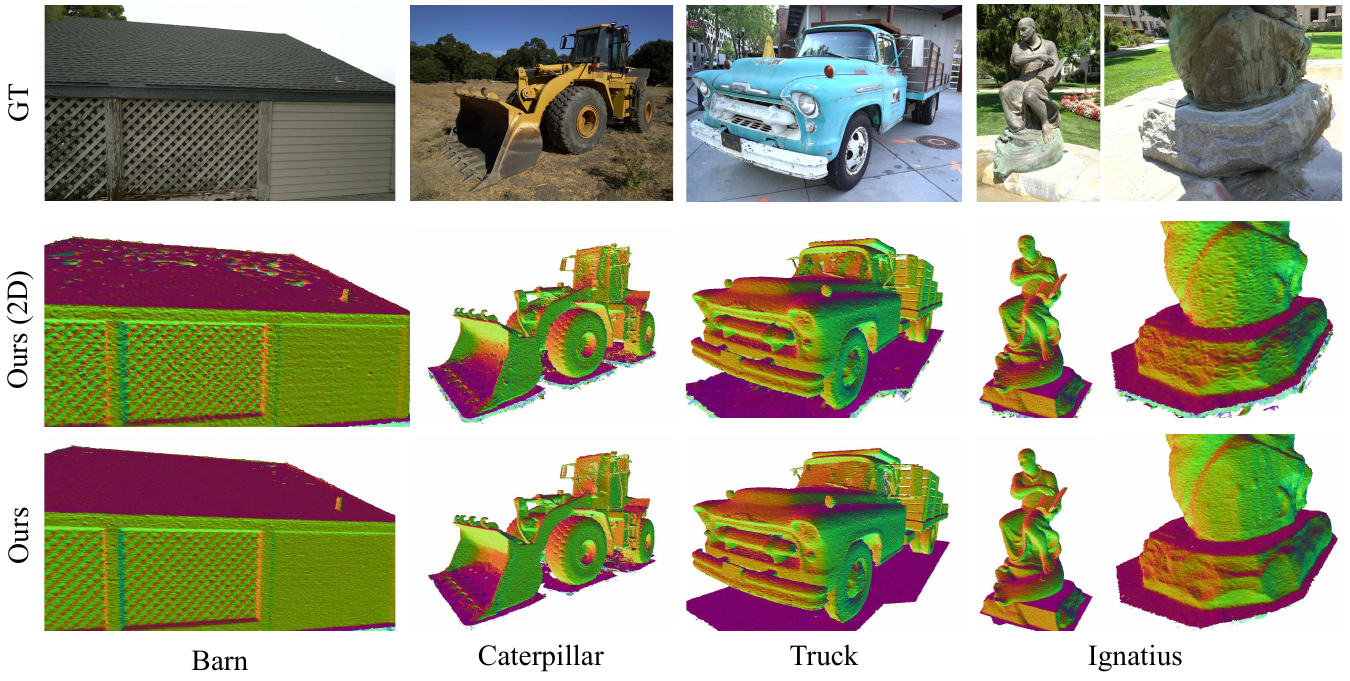}
\centering
\caption{\textbf{Mesh Extraction with Different Gaussian Rasterizer.} } 
\label{fig:supp_tnt}
% \vspace{-0.5cm}
\end{figure*}

\subsection{Ablation on Different Rasterizers}
In Figure \ref{fig:supp_tnt}, we present reconstruction results using different Gaussian rasterizers, showcasing the versatility and adaptability of our method. These results demonstrate that our approach can be seamlessly integrated into existing Gaussian methods, no matter it's 2D Gaussians \cite{huang20242d} or 3D Gaussians \cite{yu2024gaussian}, enhancing their performance without requiring significant modifications. Furthermore, our method consistently delivers high-quality reconstructions with fine details, highlighting its effectiveness in capturing intricate scene structures such as transparent and reflection areas.

% \subsection{Ablations on Normalization Function.} 
% We acknowledge that non-uniform transformation like MipNeRF's \cite{barron2022mip} non-linear normalization (mapping $\pm \infty$ to $(-2,2))$ can cause distortions and conflicts with tinycudann's requirements of having inputs in $(0, 1)$. In contrast, our method applies a near-linear transformation for Gaussians within the initial bounding box and a compression for outliers, minimizing distortion effects. As shown in Table \ref{tab:supp}, our approach (``Ours") outperforms MipNeRF's normalization function (``w/ MipNeRF NF") which is also scaled to $(0, 1)$.

% \subsection{Ablations on SDF to Opacity Function.} 
% We compare results using 3DGSR's \cite{lyu20243dgsr} SDF-to-Opacity function (denoted as ``w/ 3DGSR SDF2O" in Table \ref{tab:supp}) with our proposed function (Eq.\ 5). Our method achieves a higher f-score, offering a simpler, more efficient solution and the ability to reach full opacity of 1, unlike 3DGSR's capped opacity of 0.25. While 3DGSR targets object-level reconstruction, this limitation may impede its effectiveness in unbounded outdoor scenes.

% \input{tables/supp}

\subsection{Ablations on Geometry Cues.}
\begin{figure*}
\includegraphics[width=11.5cm]{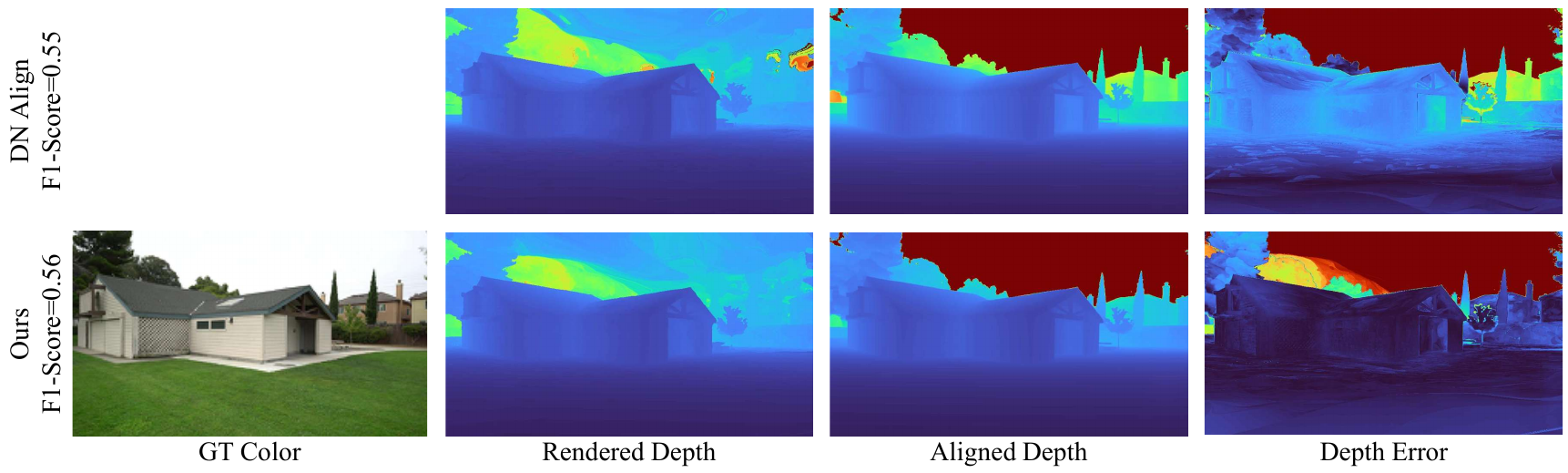}
\centering
\caption{\textbf{Ablations on Geometry Cues.} We show the ground truth color image, rendered depth, aligned pseudo depth and depth error (in log) on Barn. DN Align represents our pipeline with DN Splatter's \cite{turkulainen2024dn} alignment.} 
\label{fig:supp_ab_geo}
% \vspace{-0.5cm}
\end{figure*}

Our analysis reveals significant artifacts in Depth Anything v2's estimated depth maps due to inherent scale uncertainty, despite global alignment attempts. As shown in Figure \ref{fig:supp_ab_geo}, we compare our alignment method ``Ours" with DN Splatter's approach \cite{turkulainen2024dn} ``DN Align", demonstrating our method's superior ability to reduce depth errors between rendered and aligned pseudo depth. Interestingly, despite the inaccuracies in DN Splatter's aligned pseudo depth, the reconstruction F1 score remains comparable to our method. This finding suggests that while these limitations underscore the unreliability of pseudo depth maps for precise reconstruction, necessitating our model's strong ability to compensate for misalignments, the primary improvement stems from our novel Gaussian-SDF collaboration pipeline rather than geometric cues alone.

% TODO: show that with dn splatter aligment function, the depth error is mucha larger

\subsection{Ablations on Pruning Strategy.}
Our method links the opacity value of each Gaussian to it's SDF value with our SDF-to-Opacity function. While this maintains computational equivalence to conventional opacity-based pruning, it provides a physically-grounded criterion.

\subsection{Ablations on Multi-Resolution Regularization.}
\begin{figure*}[t]
\includegraphics[width=11.5cm]{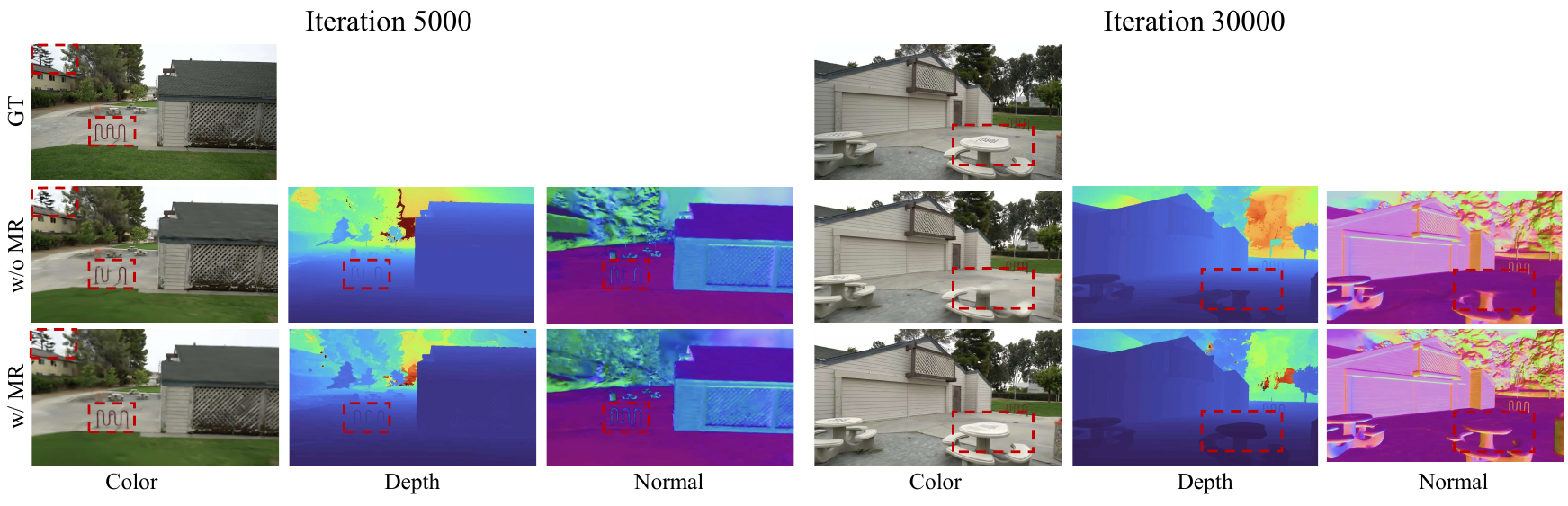}
\centering
\caption{\textbf{Ablation Study on Multi-Resolution Regularization.} We show a comparison of ground truth color images alongside our rendered color, depth, and normal outputs at various training iterations on Barn. The results demonstrate that our multi-resolution regularization enables more efficient and complete scene geometry reconstruction, achieving superior convergence compared to standard training approaches.} 
\label{fig:supp_ab_mr}
% \vspace{-0.5cm}
\end{figure*}

\begin{figure}
\includegraphics[width=5.5cm]{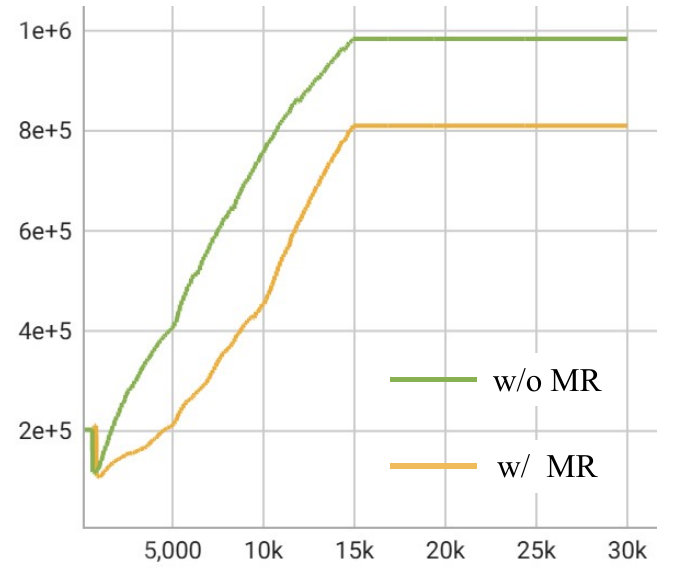}
\centering
\caption{\textbf{Total Gaussian Population Growth on Barn (Training Iterations vs. Gaussian Count).} Our proposed multi-resolution regularization achieves comparable reconstruction and rendering quality with significantly fewer Gaussians than standard training approaches, demonstrating improved computational efficiency without compromising output quality.} 
\label{fig:supp_ab_mr2}
% \vspace{-0.5cm}
\end{figure}

Our Multi-Resolution training strategy employs a coarse-to-fine approach, initially utilizing fewer Gaussians at lower resolutions to establish robust scene geometry before progressively refining details. Figure \ref{fig:supp_ab_mr2} quantitatively illustrates the controlled growth of Gaussian populations throughout training. While the quantitative results remain the same since the evaluation only focus on foreground objects, the qualitative results shows significant improvement. Figure \ref{fig:supp_ab_mr} demonstrates the effectiveness of our approach, showing that multi-resolution regularization enables reliable reconstruction of objects with limited views, whereas standard training struggles in such scenarios.

\subsection{Ablations on Different Meshing Methods.} 
\begin{figure}
\includegraphics[width=8.5cm]{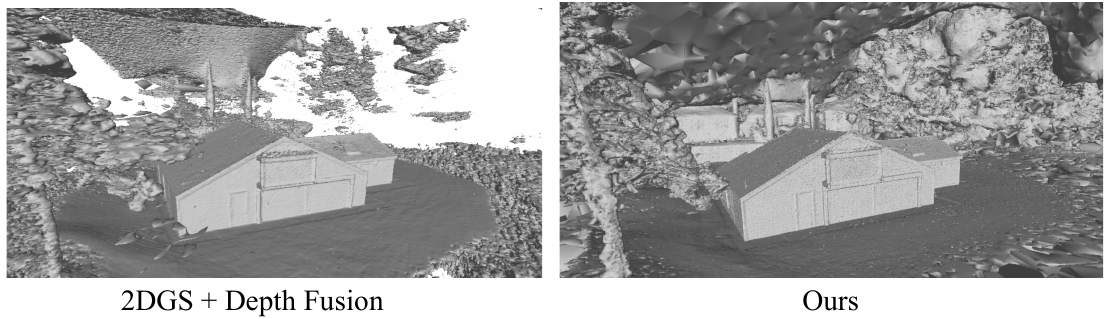}
\centering
\caption{\textbf{Comparison on Depth Fusion and Ours on Barn.}} 
\label{fig:supp_ab_meshing1}
% \vspace{-0.5cm}
\end{figure}

\begin{figure}
\includegraphics[width=8.5cm]{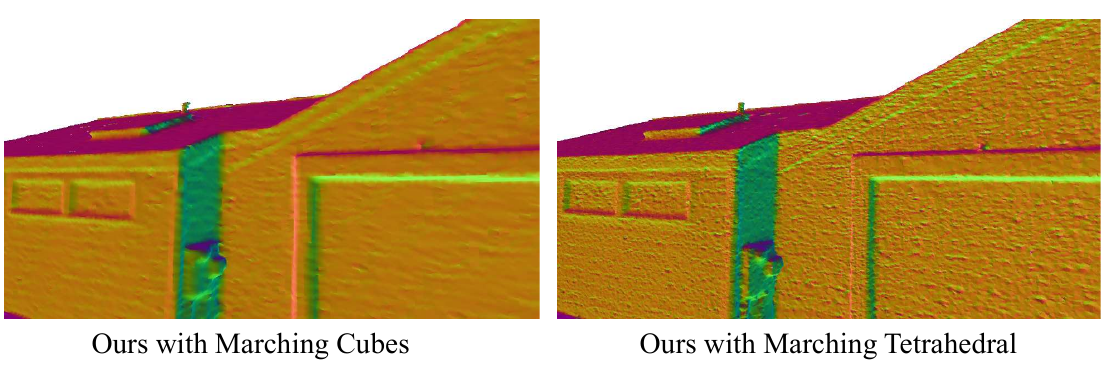}
\centering
\caption{\textbf{Ablation Study on Different Meshing Methods.} While Marching Cubes struggles with resolution limitations in unbounded scene reconstruction, resulting in low-quality meshes with missing details, our method consistently produces high-quality meshes with enhanced geometric fidelity and preserved fine structures.} 
\label{fig:supp_ab_meshing2}
% \vspace{-0.5cm}
\end{figure}

Figure \ref{fig:supp_ab_meshing1} illustrates the limitations of depth fusion-based mesh extraction in unbounded scenes, revealing significant distortion caused by resolution constraints and MipNeRF's \cite{barron2022mip} normalization artifacts. Comparative results in Figure \ref{fig:supp_ab_meshing2} demonstrate our method's advantages across different meshing techniques, showing that our Gaussian-guided surface reconstruction achieves both superior efficiency and enhanced detail preservation compared to existing approaches.

\section{Limitation Discussion}
The pseudo depth maps exhibit high uncertainty in the far distance regions and inconsistency across multiple views, which limits our method's ability to reconstruct distant background areas, such as the sky. However, since foreground objects are the primary focus in reconstruction tasks, our method demonstrates superior performance in these regions. To address this limitation, we plan to introduce an uncertainty-based weighting mechanism for the geometry regularization.

% \begin{nolinenumbers}
\input{tables/supp_dtu}

% \end{nolinenumbers}
% To split the supplementary pages from the main paper, you can use \href{https://support.apple.com/en-ca/guide/preview/prvw11793/mac#:~:text=Delete%20a%20page%20from%20a,or%20choose%20Edit%20%3E%20Delete).}{Preview (on macOS)}, \href{https://www.adobe.com/acrobat/how-to/delete-pages-from-pdf.html#:~:text=Choose%20%E2%80%9CTools%E2%80%9D%20%3E%20%E2%80%9COrganize,or%20pages%20from%20the%20file.}{Adobe Acrobat} (on all OSs), as well as \href{https://superuser.com/questions/517986/is-it-possible-to-delete-some-pages-of-a-pdf-document}{command line tools}.

%% file: tables/supp_dtu.tex
\setlength\tabcolsep{0.5em}
\begin{table*}[t]
\centering

% \vspace{-0.2cm}
\resizebox{.98\textwidth}{!}{
\begin{tabular}{@{}lcccccccccccccccclc}
\hline
 \multicolumn{2}{c}{} & 24 & 37 & 40 & 55 & 63 & 65 & 69 & 83 & 97 & 105 & 106 & 110 & 114 & 118 & 122 & & Mean \\ 
 \cline{1-19}
 N-angelo~\cite{li2023neuralangelo} & & 0.49 & 1.05 & 0.95 & 0.38 & 1.22 & 1.10 & 2.16 & 1.68 & 1.78 & 0.93 & 0.44 & 1.46 & 0.41 & 1.13 & 0.97 & & 1.07\\ 
 N-angelo*~\cite{li2023neuralangelo} & & 0.37 & 0.72 & 0.35 & 0.35 & 0.87 & 0.54 & 0.53 & 1.29 & 0.97 & 0.73 & 0.47 & 0.74 & 0.32 & 0.41 & 0.43 & & 0.61\\ 
 Ours & & 0.45 & 0.65 & 0.36 & 0.36 & 0.94 & 0.70 & 0.67 & 1.27 & 0.99 & 0.63 & 0.49 & 0.84 & 0.39 & 0.53 & 0.47 && 0.65 \\
 \hline
\end{tabular}
}
\caption{\textbf{Quantitative Comparison on the DTU Dataset~\cite{jensen2014large}}. We show the Chamfer distance. For Neuralangelo \cite{li2023neuralangelo}, we report the results from UniSDF \cite{wang2023unisdf} reproduction as N-angelo, and the results from Neuralangelo publication as N-anglo*, which is not reproducible.}
\label{tab:supp_dtu}
% \vspace{-0.1cm}
\end{table*}